\documentclass{article} 
\usepackage{times}
\usepackage{graphicx}
\usepackage{natbib} 
\usepackage{amsmath} 
\usepackage{amssymb}
\usepackage{multirow}
\usepackage{subfigure}
\usepackage{hyperref}
\usepackage{sidecap}
\usepackage{url}
\usepackage{epstopdf}
\usepackage{soul}
\usepackage[accepted]{icml2014}

\newcommand{\BEQ}{\begin{equation}}
\newcommand{\EEQ}{\end{equation}}
\newcommand{\BEA}{\begin{eqnarray}}
\newcommand{\EEA}{\end{eqnarray}}
\newcommand{\be}{\begin{eqnarray}}
\newcommand{\ee}{\end{eqnarray}}
\newcommand{\benn}{\begin{eqnarray*}}
\newcommand{\eenn}{\end{eqnarray*}}
\newcommand{\vx}{{\mathbf x}}
\newcommand{\vy}{{\mathbf y}}
\newcommand{\ii}{{(i)}}
\newcommand{\jj}{{(j)}}
\newcommand{\glass}{\textit{glass}}
\newcommand{\wine}{\textit{wine}}
\newcommand{\iris}{\textit{iris}}
\newcommand{\shortpara}[1]{{\bf #1~~~}}

\newcommand{\eat}[1]{}

\icmltitlerunning{Demystifying Information-Theoretic Clustering}

\begin{document}

\twocolumn[
\icmltitle{Demystifying Information-Theoretic Clustering}

\icmlauthor{Greg Ver Steeg$^{1,2}$}{gregv@isi.edu}
\icmlauthor{Aram Galstyan$^{1,2}$}{galstyan@isi.edu}
\icmlauthor{Fei Sha$^{2}$}{feisha@usc.edu}
\icmlauthor{Simon DeDeo$^{3,4}$}{simon.dedeo@gmail.com}
\icmladdress{
$^{1}$ Information Sciences Institute, 4676 Admiralty Way, Marina del Rey, CA 90292, USA \\
$^{2}$ University of Southern California, Los Angeles, CA 90089, USA \\
$^{3}$ Santa Fe Institute, 1399 Hyde Park Rd., Santa Fe, NM 87501, USA \\
$^{4}$ School of Informatics and Computing, Indiana University, 901 E 10th St., Bloomington, IN 47408, USA}

\icmlkeywords{clustering, information theory}

\vskip 0.3in
]

\begin{abstract} 

We propose a novel method for clustering data which is grounded in information-theoretic principles and requires no parametric assumptions. 
Previous attempts to use information theory to define clusters in an assumption-free way are based on maximizing mutual information between data and cluster labels.
We demonstrate that this intuition suffers from a fundamental conceptual flaw that causes clustering performance to deteriorate as the amount of data increases.
Instead, we return to the axiomatic foundations of information theory to define a meaningful clustering measure based on the notion of consistency under coarse-graining for finite data. 
\end{abstract} 
 
\section{Introduction} 

Clustering is a fundamental problem in machine learning~\cite{clusterreview}.  
As the objective of clustering is typically exploratory in nature, we desire clustering algorithms that make as few assumptions about the data as possible. We would like those algorithms to be flexible enough to reveal complex data patterns that are 
nonlinear, multi-modal and
invariant with respect to changes in data representation. Ideally, we would like to achieve those goals without 
explicitly defining some notion of similarity between data points~\cite{slonim} or defining ``prototypical'' clusters~\cite{ric}.

Latent variable models are a common approach to clustering. This is exemplified by the wide adoption of Gaussian mixture models and its simplified version $k$-means.  
While maximizing likelihood of the data under some probabilistic model is a clear and operationally meaningful criteria, such models are not invariant to changes of data representation and require a fully specified generative model.

Ostensibly, information-theoretic criteria satisfy all our desiderata. There has been a growing interest in information-theoretic clustering  where we assign cluster labels to data points such that the mutual information between data and labels is maximized~\cite{faivishevsky,feicluster,spanning,squaredloss}. Mutual information captures higher-order statistics in the data and is suitable for complex data patterns where linear approaches or Gaussian assumptions are inadequate. Moreover, mutual information is invariant with respect to  smooth, invertible transformations of data~\cite{cover}. It can also be estimated non-parametrically from the data samples without defining a parametric space of probability distributions~\cite{kraskov}. This is especially attractive for clustering high-dimensional data.

The main contribution of this paper is to demonstrate that information-theoretic clustering based on mutual information is fundamentally flawed and to derive a novel alternative grounded in information-theoretic principles. We derive a non-parametric estimator for our clustering objective and a tractable heuristic to optimize it. We demonstrate simple scenarios that cause mutual information clustering to fail dramatically while our method succeeds. 

Mutual information is naturally interpreted as a measure of compressibility of data. However, {\em compression alone does not capture natural cluster structure in the data}. 
As we illustrate with an exemplar problem, with more information about the underlying probability distribution, mutual information clustering will prefer to split the probability mass into arbitrary but equal-sized masses of probability, completely ignoring the data's intrinsic structure!
\emph{How do we reconcile this observation with the many previously reported empirical successes of information-theoretic clustering}?  We argue that the good clustering performance demonstrated by those methods 
is not due to the objective but rather reflect transient effects of the estimators used.
Thus, counterintuitively, all those methods eventually fail when given {\em more} data. In the large data limit, the estimators converge to their true values yielding equal-sized, but meaningless, clusters.

We show that a fix cannot be constructed by simply tweaking the information-theoretic objective.
Instead, we construct an objective from first principles that preserves information theory axioms even when applied to finite samples from an unknown distribution. 
We motivate our approach by appeal to the axiom of consistency under coarse-graining that forms the definition of entropy~\cite{shannon}. While the consistency axiom is preserved exactly in the limit of infinite data, empirical estimation will generically lead to its violation. We shall show how larger violations signal a non-robust partition. A lower violation of consistency under coarse-graining is an important property preserved by natural cluster structures, but not by the spurious equal-sized partitions implied by mutual information.
We thus propose data be clustered such that consistency is violated minimally. 
We construct a non-parametric estimator for this quantity and demonstrate an alternate interpretation of consistency violation as \emph{cluster label uncertainty}. 

We validate the proposed approach on synthetic data and commonly used datasets for clustering and contrast to existing approaches for information-theoretic clustering. For synthetic data, we show that our measure overcomes the shortcomings of previous information-theoretic estimators, recovering natural clusters even in the limit of large data. We also introduce a heuristic to optimize this objective and we show that it recovers non-convex clusters and achieves competitive clustering results on standard datasets. 

The rest of the paper is organized as follows. In Sec.~\ref{sec:problem}, we describe the basic idea of information-theoretic clustering and point out the pitfalls of the status quo which equates compression with clustering.  In Sec.~\ref{sec:coarse}, we describe the proposed approach, starting by developing the idea of coarse-graining. In Sec.~\ref{sec:results}, we report empirical studies on applying our approach to synthetic and real-world datasets. Related work, conclusions, and future research directions are discussed in Sec.~\ref{sec:discussion}.



\section{Information-theoretic clustering and its pitfalls}\label{sec:problem}
Given samples drawn i.i.d. from a {\em known} distribution, the Shannon entropy of the distribution can be interpreted as the minimum number of bits needed to encode each sample (on average). For clustering, we are given samples of an {\em unknown} distribution, and we would like to label (encode) each sample to reflect some natural structure. Even if we knew the Shannon entropy of the distribution, a code that achieves this optimal compression
does not necessarily reflect the natural structure of the underlying distribution.

\subsection{Basic concepts and entropy estimation}\label{sec:basic}

 We begin with the generic clustering problem in which we are given some samples $\vx^\ii \in \mathbb R^d$ for $i=1,\ldots,N$, drawn from some unknown distribution, $p(\vx)$. The goal is to associate a discrete label, $y^\ii \in \{0,1\}$ (for simplicity we use only binary labels), that somehow reflects a {\em natural} clustering of the data. Of course, the main difficulty is in defining what qualifies as a natural clustering. In what follows, we consider various information-theoretic criteria.

Entropy is defined in the usual way as $H(X) = \mathbb E [ -\log p(\vx)]$. We use base-two logarithms so that information is measured in bits. Using standard notation, capital $X$ denotes a random variable whose instances are denoted in lowercase, and the fact that entropy is a functional of the probability distribution is only explicitly written when clarity demands it.  The expectation value may be a sum for discrete variables or an integral in the continuous case. Higher-order entropies can be constructed in various ways from this standard definition. For reference, we provide a few alternate forms of the mutual information, $I(X;Y)$.
\benn
 I(X;Y) &=& H(X)+H(Y)-H(X,Y) \\
 &=& H(X) - H(X|Y) \\
 &=& H(Y)-H(Y|X)
 \eenn
 
A function that estimates entropy from some i.i.d. samples drawn from $p(\vx)$ we denote with $\hat H(X)$.  An intuitive way to estimate entropy directly from samples in a non-parametric way follows.
\be\label{eq:kraskov}
H(X)  = \mathbb E ( \log \frac{1}{p(\vx)})  &\approx&  \frac{1}{N} \sum_{i=1}^N \log \frac{1}{p(\vx^\ii)} \\
 & \approx &  \frac{1}{N} \sum_{i=1}^N \log \frac{\epsilon_{i,k}^d}{k/N}  \nonumber
 \ee
 \vspace{-5mm}
 \be
 \hat H(X)  \equiv  \log (N/k) + \frac{d}{N} \sum_{i=1}^N \log \epsilon_{i,k} + c_{k,N} 
\ee
In the first line, we take the sample mean instead of the expectation value, but we still have the (unknown) density, $p(\vx^\ii)$, in the expression. On the next line, we locally approximate this density by making the smallest box that contains $k$ neighboring points. Then the density is approximated by $k/N$, the fraction of points in the box, over the volume, $\epsilon_{i,k}^d$, where $\epsilon_{i,k}$ denotes the distance to the $k$-th nearest neighbor of point $\vx^\ii$ according to the max-norm. 
In our definition in the second line, we add a small constant factor $c_{k,N} =\psi(N)-\psi(k)+\log(2k/N)$ to match with the non-parametric, asymptotically unbiased Kozachenko-Leonenko entropy estimator \cite{Kozachenko} (or k-nearest neighbor, or kNN, estimator), which requires a more involved derivation as provided by Kraskov et. al.~\cite{kraskov}. 
We write it in this alternate format to increase intuition and to ease later derivations.
Note that this estimator depends only on distances between neighboring data points. The estimator has also been empirically shown to have good performance for small amounts of data, with $k=3$ representing a good choice~\cite{kraskov,empiricalmi}. 
For discrete variables, we use the standard plug-in estimator. DeDeo et. al. discuss more nuanced alternatives in the discrete case~\cite{dedeo}.

\subsection{Pitfalls}

\begin{figure}[ht] 
\centering
    \hspace{-5mm}\includegraphics[width=0.93 \columnwidth]{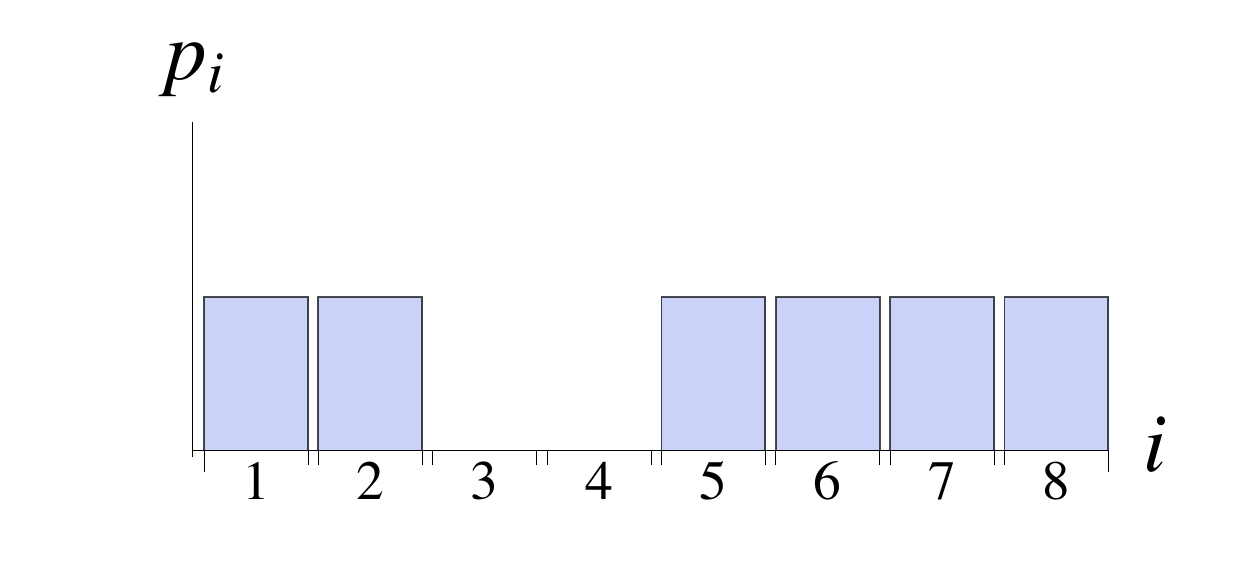} \\ (a) \\
     \includegraphics[width=0.85 \columnwidth]{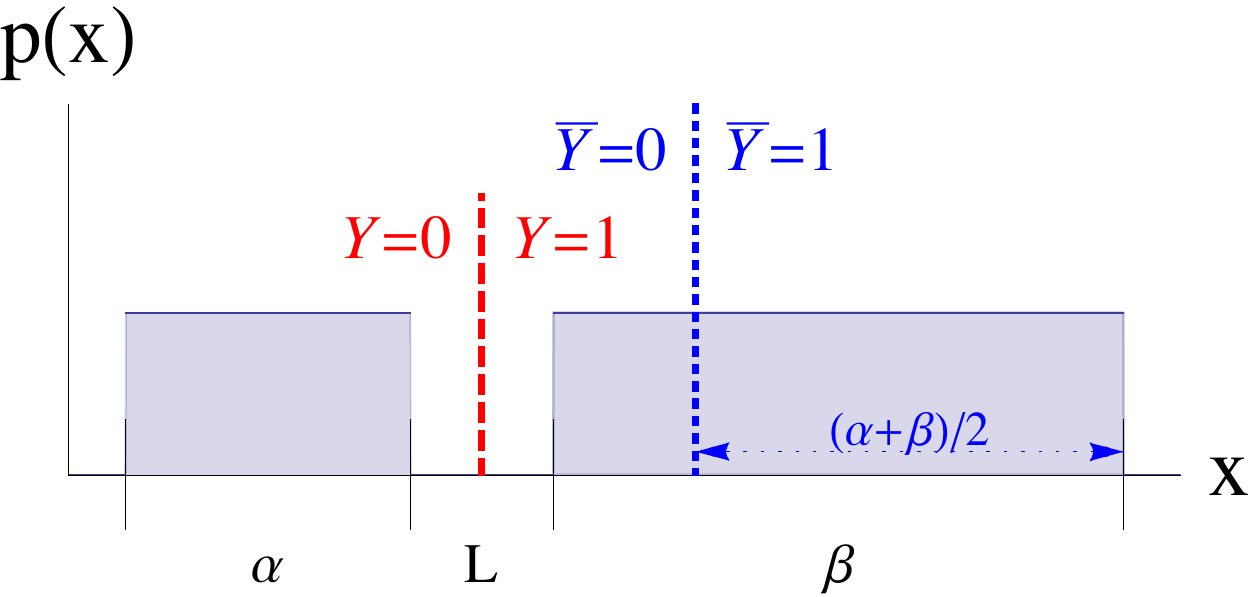} \\ (b)
   \caption{(a) A discrete clustering problem. Because the bin numbers are arbitrary, no purely information-theoretic criteria can recover the intuitive clustering that splits the bins based on the gap. (b) The continuous version can be viewed as the limiting case of the discrete picture above with infinitely precise bin widths. The ``clustering'' which maximizes the mutual information between cluster labels and $x$ is denoted with $\bar Y$, leading to $I(X;\bar Y) = 1.$ }
   \label{fig:bar}
      \vskip -0.2in
\end{figure}

\shortpara{Compression $\ne$ clustering}  A frequently invoked and plausible sounding intuition is that we should maximize mutual information between data and cluster labels. 
Consider first the simple case of clustering discretely distributed data, exemplified in Fig.~\ref{fig:bar}(a).  Is there a purely information-theoretic criteria that clusters the bins of this discrete probability distribution into the two natural groups separated by the gap? There cannot be, because from a purely information-theoretic perspective the bin labels are arbitrary. That is, the bins can be re-ordered arbitrarily so that, e.g., there is no gap, and this re-ordering does not affect any information-theoretic quantity because they depend only on the values $p_i$. While no one would attempt to cluster these categorical variables without defining some relationship between the variables, the same issue arises in a more subtle form for continuous distributions.

In the simplest picture of clustering for continuous variables, we have a mixture of two uniform probability distributions in one dimension, shown in Fig.~\ref{fig:bar}(b). If we split the two pieces according to the intuitive clustering, a simple analytic calculation gives $I(X;Y) = H_0 (\alpha/(\alpha+\beta))$, where $H_0$ represents the binary entropy function which is in the range $[0,1]$. If we split the space into two equally sized masses of probability, we maximize the mutual information, $I(X;\bar Y)=1$. Clearly, maximizing mutual information does not have the intended effect. Even slightly unbalanced clusters will not be found.
Essentially, the scenario in Fig.~\ref{fig:bar}(b) is the limit of the case in Fig.~\ref{fig:bar}(a) with an infinite number of bins that are infinitely narrow~\cite{cover}.
These infinitesimal bins can be re-ordered arbitrarily without affecting the value of any information-theoretic (IT) measure.

\shortpara{The mystery about the empirical success} Although this example makes it clear that a good clustering is not the one that maximizes mutual information, or any solely IT criteria, this leads to a mystery. How have so many papers achieved good clustering performance using this criteria~\cite{feicluster,faivishevsky,spanning}? To understand this result, it helps to write mutual information in the form $I(X;Y) = H(Y) - H(Y|X)$. The first term, $H(Y)$ is maximized for equally sized clusters. The second term, $H(Y|X)$, should be 0 for any exact partitioning of the input space (in which case $y=f(\vx)$). 

\begin{figure}[ht] 
\centering
    \hspace{-9mm}\includegraphics[width=0.83 \columnwidth]{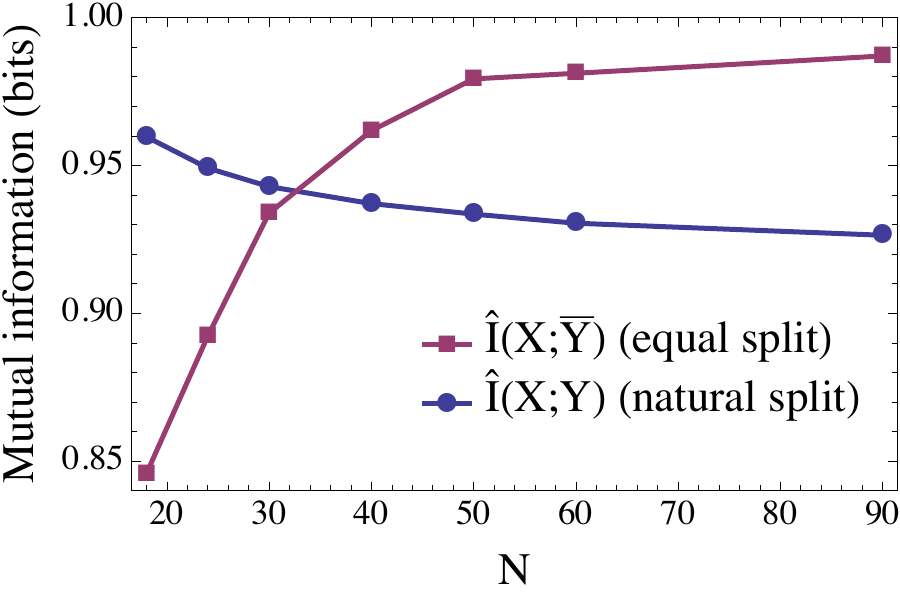} \\ (a) \\
    \includegraphics[width=0.85 \columnwidth]{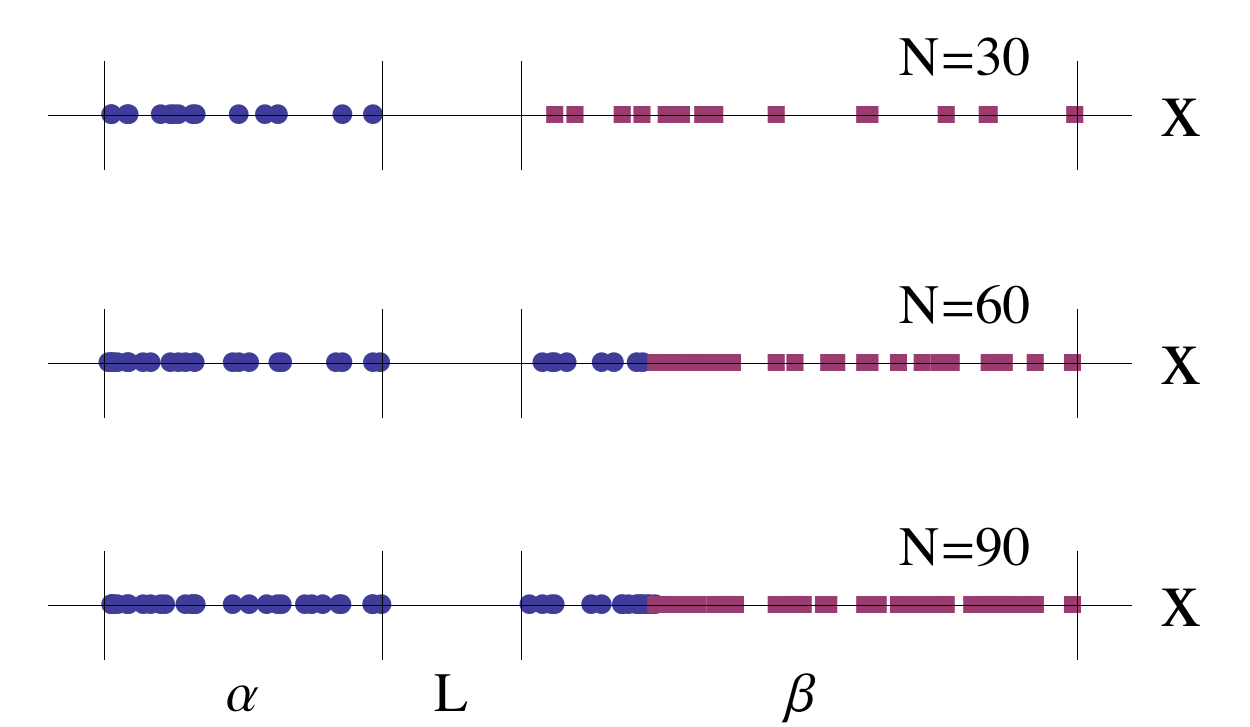} \\ (b)
   \caption{(a) We compare the estimated mutual information of the ``natural'' clustering with the one that splits the probability mass equally. We used the distribution in Fig.~\ref{fig:bar}(b) with $\alpha=1,L=0.5,\beta=2$. As $N$, the number of samples, increases, eventually the equal split becomes the preferred one (maximizing estimated mutual information). (b) Clustered samples at various $N$.}
   \label{fig:sample}
\end{figure}

However, it is easy to see that if we cluster a finite sample of data points, as in Fig.~\ref{fig:sample}(b), that our {\em estimate}, $\hat H(Y|X)$, will not be zero (using, e.g., the non-parametric estimator introduced in Sec.~\ref{sec:basic}). In particular, this will be the case near the boundary between the two clusters. The natural clustering will have a smaller value for $\hat H(Y|X)$ (due to the gap of width $L$ separating the clusters). On the other hand, $\hat H(Y)$ will be larger for clusters of equal size. These two terms compete. For small amounts of data, we can see in Fig.~\ref{fig:sample}(a) that the natural clustering is preferred, i.e. has higher estimated mutual information. However, as the amount of data increases, the uncertainty decreases and eventually equal-sized clusters will be preferred. Ironically, more data leads to a less desirable result. 
This behavior does not depend on the estimator used: any estimator that is asymptotically unbiased will converge to the same value in the large $N$ limit.
The behavior also persists for arbitrary distributions because the contribution of $\hat H(Y|X)$ comes from points near the boundary. For any clusters, the percentage of points near the boundary will decrease as $N$ increases. Tests with previous information-theoretic clustering objectives focused on small, nearly balanced datasets (like many of the UCI datasets considered in Sec.~\ref{sec:results}), so that these shortcomings went unnoticed.

Precision is both the problem and the solution. Increasing $N$ reduces the uncertainty in cluster labels near the cluster boundaries, so that any clustering has approximately the same value of $\hat H(Y|X)$. 
On the other hand, given a sample of $N$ points, we can always resample to include fewer points. Then we can test whether a clustering is more natural in the sense that it reduces the uncertainty in the cluster labels, even for small amounts of data.
We make this intuition precise in the next section.

\section{Proposed approach}\label{sec:coarse}

We saw that previous information-theoretic clustering schemes only work accidentally insofar as they mis-estimate the true uncertainty of cluster labels near the cluster boundaries because of finite data. This behavior is closely linked to the axioms of information theory. We briefly make this connection before deriving a simple expression that explicitly makes use of the uncertainty near cluster boundaries due to limited data. 

\subsection{Shannon's Axiom of Consistency under Coarse-Graining}

Shannon's original derivation of entropy~\cite{shannon} begins with properties that a measure of uncertainty should be expected to obey and concludes that (Shannon) entropy is the unique measure that satisfies these properties. Consider a discrete probability distribution where outcome $j$ occurs with probability $p_j \equiv p(y=j)$. The first two desired properties are continuity (a small change in $p_j$ does not cause a large change in uncertainty) and that uncertainty should grow with $k$ if there are $k$ equally likely events. 
The final property we refer to as {\em consistency under coarse-graining}.
\benn
\lefteqn{h(p_1,p_2,p_3) = h(p_1,p_2+p_3) + } \\
 && (p_2+p_3)~ h(p_2/(p_2+p_3),p_3/(p_2+p_3)).
\eenn
Intuitively, the measure of uncertainty should not change if we lump together (coarse-grain) bins 2 and 3. After combining the uncertainty of the coarse-grained bins with the (weighted) uncertainty within each coarse-grained bin we should recover the original measure of uncertainty. DeDeo et. al. make the further point that any {\em estimator} of entropy should at least approximately satisfy this property or it risks losing its meaning as an information-theoretic measure altogether~\cite{dedeo}. 

What is the natural analogue of the coarse-graining criteria for continuous distributions? Suppose we split our space into two disjoint regions $\mathcal R_0, \mathcal R_1$ so that $ \mathcal R_1  = \mathbb R^d \setminus \mathcal R_0$. For some continuous distribution $p(\vx)$, we can define a joint probability $p(\vx,y)$, where $y=j \leftrightarrow \vx \in \mathcal R_j,$ or equivalently,  $y=f(\vx), f: \mathbb R^d \rightarrow \{0,1\}$. Then the probability of drawing a point in region $j$ is just 
$\int_{\mathcal R_j}~dx ~p(\vx) = \int ~dx~p(\vx, y=j) = p(y=j).$
Consistency under coarse-graining would demand,
\be\label{eq:consistency}
\lefteqn{H(p(\vx)) = H(p(y)) +} \\
&& p(y=0) H(p(\vx | y=0))+ p(y=1) H(p(\vx| y=1)), \nonumber
\ee
which is easily shown to be satisfied for differential entropy. 
This can be written in the more succinct, standard notation as $H(X) = H(Y)+H(X|Y)$\footnote{Note that $H(X|Y) \equiv \sum_i p(y=i) H(p(x|y=i))$, and that this condition should only hold if $y=f(\vx)$}.


Following the logic for discrete entropy estimators, we would like to check if differential entropy estimators based on finite data obey some notion of consistency under coarse-graining. 
It is easily shown that no estimator can satisfy this condition for {\em arbitrary} coarse-grainings and remain asymptotically unbiased.\footnote{Imagine we have $N$ samples from an unknown distribution, $p(\vx)$. We can randomly partition the $N$ samples into two equal size sets and define regions $\mathcal R_0, \mathcal R_1$ accordingly to contain both sets. Call the entropy estimates for all $N$ points and for a random subset of $N/2$ points $\hat H_N(X),\hat H_{N/2}(X)$, respectively. In the large $N$ limit, we expect both of these quantities to converge to the true entropy for an asymptotically unbiased estimator. On the other hand, consistency (Eq.~\ref{eq:consistency}) would require $\hat H_N(X) = 1+\hat H_{N/2}(X)$, a clear conflict.}
Since we cannot construct an asymptotically unbiased entropy estimator that is consistent under arbitrary coarse-graining, instead we start with an unbiased estimator and search for coarse-grainings that lead to consistent entropy estimates. We refer to these coarse-grainings as {\em natural}. 
In principle, this argument can be applied to any entropy estimator, but we choose to focus on non-parametric estimators in keeping with our goal to minimize assumptions about the data. 

\begin{figure}
\centering
\includegraphics[height=0.7in]{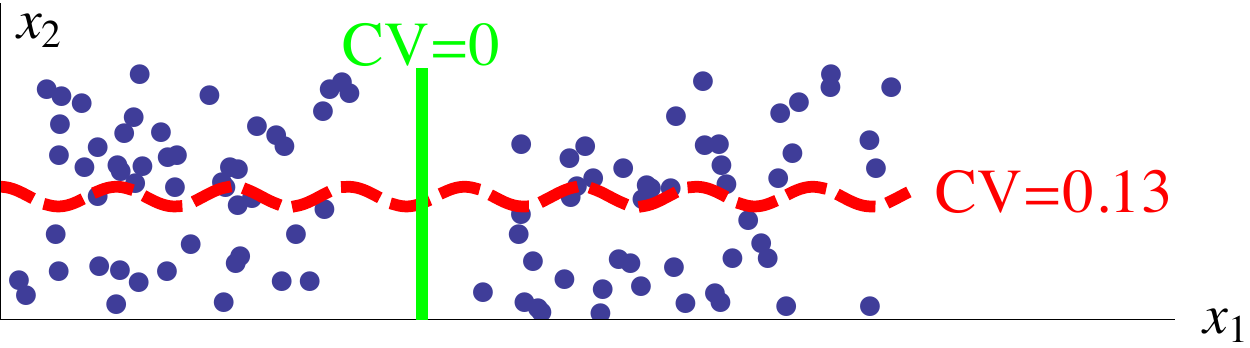}
\caption{Splitting the space using the dashed line is one (unnatural) way of coarse-graining. The solid line represents a coarse-graining that is more natural as quantified in Eq.~\ref{eq:cv1}.}\label{fig:coarse}
\vskip -0.2in
\end{figure}

Given samples, $\vx^\ii$, from a continuous distribution, $p(\vx)$, 
a coarse-graining (clustering) can be defined by associating a discrete label $y^\ii$ with each sample point. 
We can then quantify the {\em consistency violation} (CV) 
with respect to Eq.~\ref{eq:consistency}, where $CV=0$ if the equality is exactly satisfied. 
\be\label{eq:CV}
CV =  \hat H(Y)+ \hat H (X|Y) - \hat H(X) 
\ee
Here $\hat H(X), \hat H(X|Y)$ are just defined by a differential entropy estimator like the one in Eq.~\ref{eq:kraskov}. 
Fig.~\ref{fig:coarse} gives an example of two different ways of coarse-graining the same data. 
The {\em natural} coarse-graining which separates well-defined clusters (solid line) produces a small consistency violation, CV, while the alternate coarse-graining (dashed line) leads to large violations. 
CV can be viewed as a measure of how well we can estimate the global entropy given the entropy of clusters of data points (i.e. the coarse-grained data). For well-separated clusters, the global entropy is just a weighted average of the cluster entropies. 
This observation provides some extra theoretical motivation for the clustering objective we propose next. 


The right-hand side of Eq.~\ref{eq:CV} looks familiar as an 
alternate form for writing $H(Y|X)$. We can simply define an estimator for the uncertainty of the cluster label given a sample point, $\hat H(Y|X)$, in terms of our previously defined estimators:
$
\hat H(Y|X)  \equiv \hat H(Y)+ \hat H (X|Y) - \hat H(X).
$
Our estimate of the uncertainty of cluster labels is exactly the same as the violation of the coarse-graining axiom due to limited data. This gives us two interpretations of our clustering objective. We want our coarse-graining to be natural in the sense that we do not violate information-theoretic axioms, even for small amounts of data. Equivalently, we want our estimated uncertainty about cluster labels to be as low as possible, even for small amounts of data. 


\subsection{Using conditional entropy for clustering}

\shortpara{Nonparametric estimation of conditional entropy} We first derive a more compelling form for Eq.~\ref{eq:CV} in Sec.~\ref{sec:derive} of the supplementary material. For each sample point, $\vx^\ii$, we associate a discrete cluster label, $y^\ii$, then, 
\be\label{eq:cv1}
\hat H_k(Y|X) = \frac{d}{N} \sum_{i=1}^N \log \frac{\bar \epsilon_{i,k}}{\epsilon_{i,k}},
\ee
where $\epsilon_{i,k}$ represents the distance to the $k$-th nearest neighbor while $\bar \epsilon_{i,k}$ is the distance to the $k$-nn restricted to points in the same cluster as sample $i$.\footnote{An ambiguity in this definition arises if there are $k$ or fewer points in a cluster. Let $n_y$ represent the number of points with the cluster label $y$. Then if $k > n_{y^\ii}+1$ we define $\bar \epsilon_{i,k} =  \epsilon_{i,N-1}$. This definition reflects the fact that our uncertainty is maximal if we are not given sufficient data. See Appendix~\ref{sec:derive} for more details.}
As long as each point's $k$ nearest neighbors lie within the same cluster, the consistency violation, or estimated uncertainty about the cluster label, will be zero, as in Fig.~\ref{fig:coarse} where we used $k=3$. 

\shortpara{Total cluster label uncertainty} For any fixed partitioning of the space, CV approaches zero for large $N$. We want our cluster label uncertainty to be small even under arbitrary resampling with limited data, so we now imagine that each point is kept with random, independent probability $1-\alpha$. We can estimate the expected value of Eq.~\ref{eq:cv1}  for any $\alpha$, $\hat H_{\alpha,k}(Y|X) \equiv \mathbb E_\alpha [\hat H_k(Y|X)]$, where the expectation is over all possible resamplings. Toward this end, we note that a point's $k$-th nearest neighbor in the resampled dataset corresponds to its $m$-th nearest neighbor ($m \ge k$)  in the original dataset with probability $q(m|k) = C_{m-1}^{k-1}(1-\alpha)^k \alpha^{m-k}$, yielding,
\be\label{eq:cv2}
\lefteqn{\hat H_{\alpha,k}(Y|X) =}\\
&& \frac{d}{N}  \sum_{i=1}^{N}\sum_{m=k}^{N-1} C_{m-1}^{k-1}(1-\alpha)^k \alpha^{m-k} \log \frac{\bar \epsilon_{i,m}}{\epsilon_{i,m}}. \nonumber
\ee
If we want $\hat H_{\alpha,k}(Y|X)$ to be small for all $\alpha$, we can just consider its value integrated over all $\alpha$. We refer to this quantity as the {\em total cluster label uncertainty} or {\em total consistency violation}. After performing an elementary integration and changing the summation variable we obtain
\be\label{eq:cv}
\lefteqn{\hat H_{T,k}(Y|X) = \int_0^1 d\alpha~\hat H_{\alpha,k} (Y|X) = } \\
&&\frac{d}{N}  \sum_{i=1}^{N}\sum_{l=1}^{N-k} \frac{k}{(l+k)(l+k-1)} \log \frac{\bar \epsilon_{i,l+k-1}}{\epsilon_{i,l+k-1}}. \nonumber
\ee
For entropy estimators, $k=1$ leads to the lowest bias but higher values of $k$ are often chosen to reduce variance~\cite{empiricalmi}. Because averaging over resamplings already reduces variance, we choose $k=1$, simplifying the expression further, and we will refer to this quantity succinctly as $\hat H_{T}(Y|X).$\footnote{For $k=1$, the log-distance of the $l$-th nearest neighbor is weighted by a factor of $1/(l(l+1))$. For comparison, the NIC objective~\cite{faivishevsky} weights all the log-distances equally. See Sec.~\ref{sec:related} for a more detailed comparison. }

Although this expression looks simple, $\epsilon_{i,l}$ is a function of all the $\vx^\jj$ and $\bar \epsilon_{i,l}$ is a function of all the $\vx^\jj, y^\jj$.
In principle, this quantity should vary between 0 (a perfect clustering) and $H(Y)$ (a completely random clustering). We can search for partitions that minimize the ratio of $\hat H_{T}(Y|X)/\hat H(Y)$ so that we can objectively compare clusterings on a scale of zero to one.
We will see in the examples that natural partitions have low CV while most partitions have a CV orders of magnitude larger. Unlike mutual information estimators, this quantity distinguishes natural cluster structure even in the large $N$ limit.



\shortpara{Numerical procedure} While the focus of this work is on deriving a principled approach to information-theoretic clustering, we should briefly mention some practical concerns. Optimizing Eq.~\ref{eq:cv} over all possible partitions is a difficult problem. 
We consider a heuristic approach which involves solving a tractable semidefinite program in Appendix~\ref{sec:sdp} of the supplementary material. We note that other information-theoretic approaches also require heuristic approaches to optimize~\cite{feicluster}.
We leave a detailed exploration of the best heuristics for this optimization, along with extensive comparisons to other clustering objectives, for future work. In the next section, we consider simple clustering scenarios where we can compare all partitions to develop intuition about the meaning of this objective. We briefly compare results from our heuristic solver to other  clustering methods for some standard datasets.

\section{Results}\label{sec:results}

\subsection{Synthetic datasets}\label{sec:synthetic}
\begin{figure}[htb] 
\centering
\subfigure[]{
    \includegraphics[width=0.47\columnwidth]{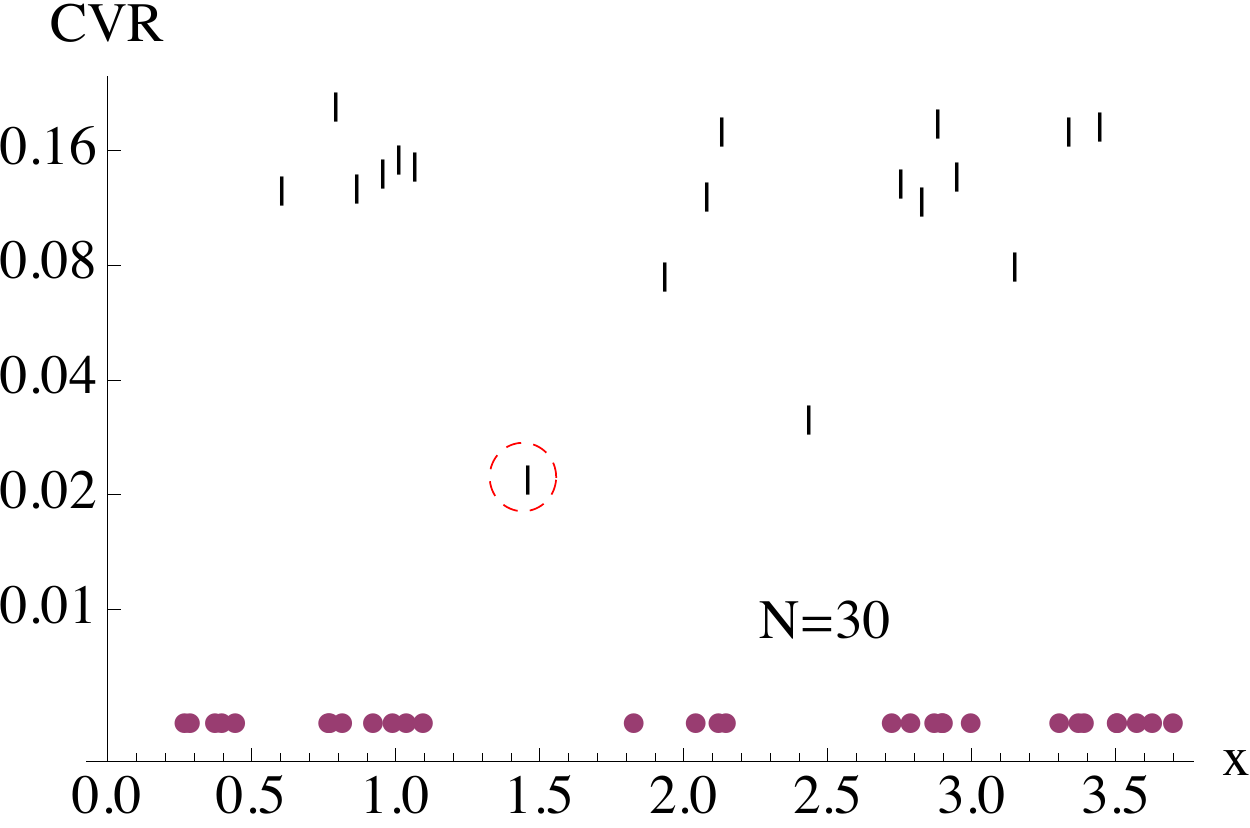} }
\subfigure[]{
    \includegraphics[width=0.47\columnwidth]{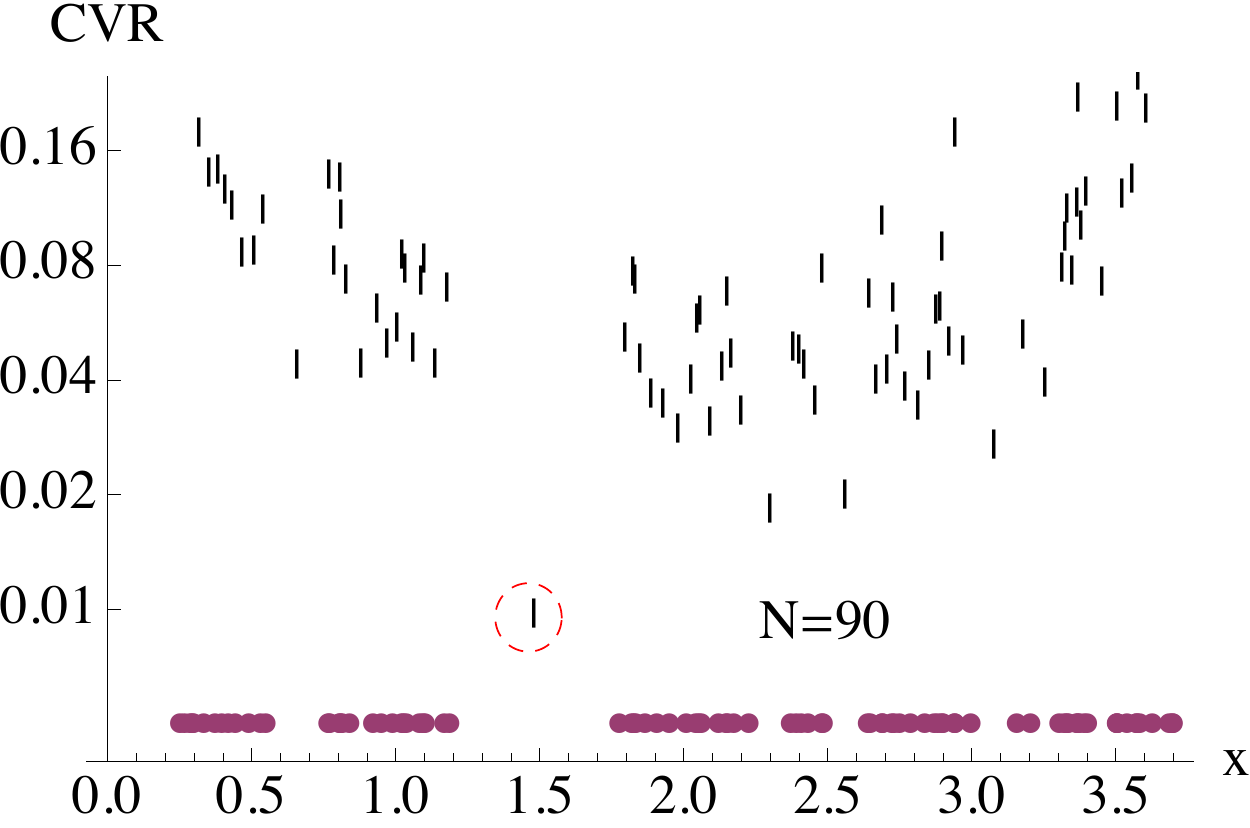} } 
   \caption{The vertical bars measure the consistency violation ratio (CVR) for partitioning the data at each value of $x$. The samples, $x^\ii$, are plotted near the $x$ axis. In this case, $\alpha=1,\beta=2,L=0.5$ (see Fig.~\ref{fig:sample}(b)) and (a)$N=30$, (b)$N=90$. Unlike the mutual information objective in Fig.~\ref{fig:sample}, adding more data does not make the clusters harder to distinguish. }
   \label{fig:cvt}
\vskip -0.2in
\end{figure}

Our  goal is to find clusterings, determined by $y^\ii$, that optimize the objective in Eq.~\ref{eq:cv}. 
Recall the simple example in Sec.~\ref{sec:problem} which mutual information failed to cluster correctly for large $N$. For data taken from the simple one-dimensional distribution in Fig.~\ref{fig:sample}, we can just measure the ratio $\hat H_T(Y|X)/\hat H(Y)$ (CVR for short from here on) for all possible ways of splitting the $x$ axis, shown in Fig.~\ref{fig:cvt}. 
We discard the trivial coarse-graining where all points are in the same group which has an undefined CVR of $0/0$, and then the best partition exactly corresponds to our intuition of the most natural clustering. We also see from Fig.~\ref{fig:cvt} that more data leads to a better separation between good and bad clusterings, as desired. 

\begin{SCfigure}[0.8][ht]
\centering
\includegraphics[width=0.45\columnwidth]{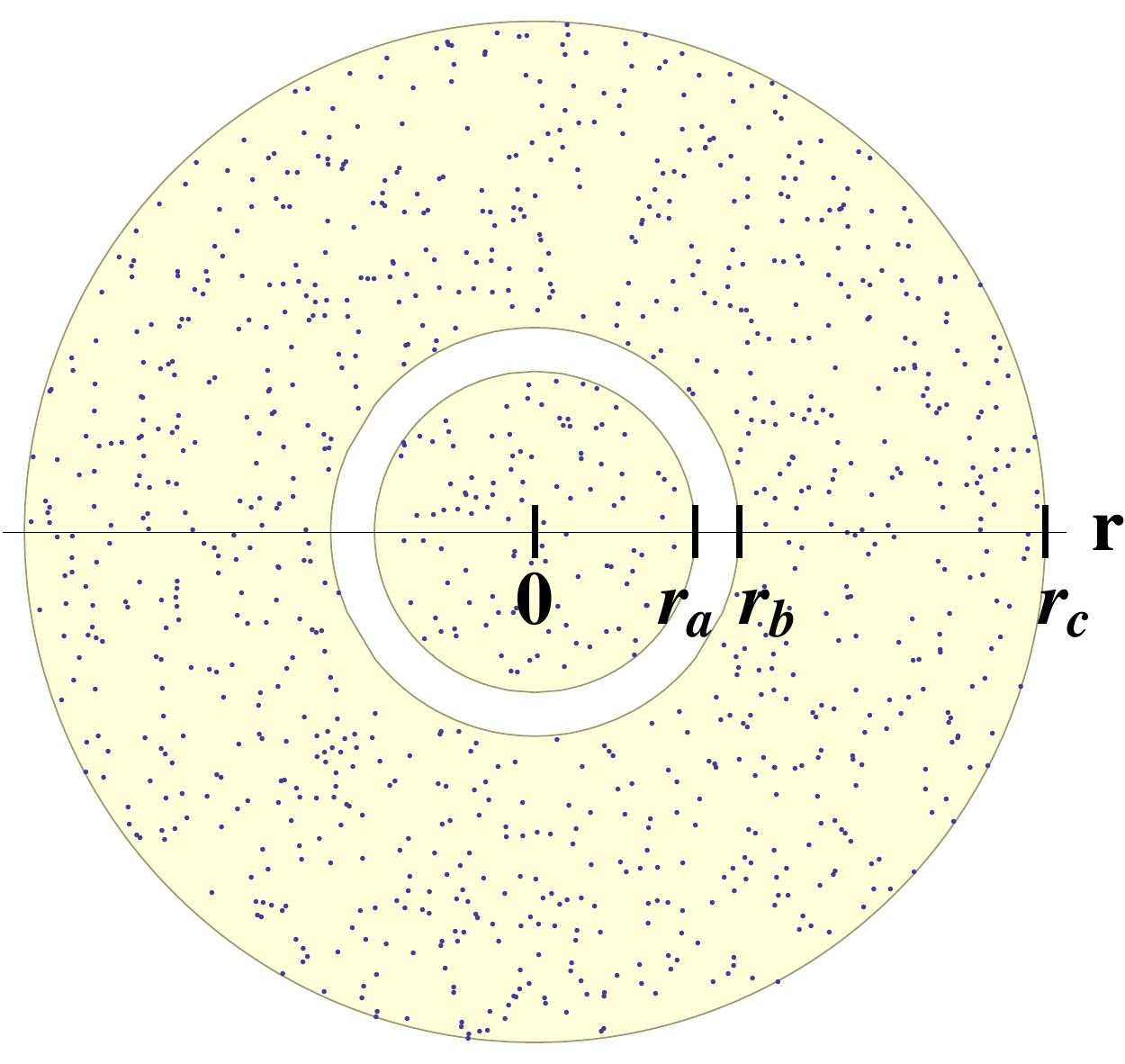}
\caption{$N$ points are drawn from the uniform distribution represented by the shaded area.}
\label{fig:circle}
\end{SCfigure}

\begin{figure}[htb]
\begin{center}
\centerline{\includegraphics[width=0.88\columnwidth]{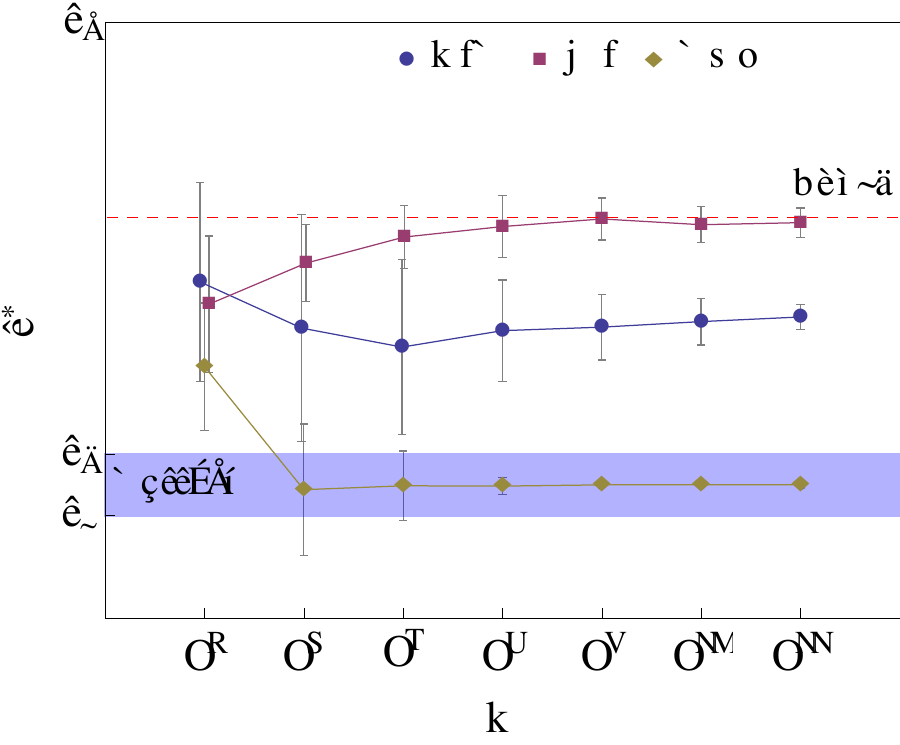}}
\caption{The ideal radius to partition the data is plotted for several information-theoretic criteria as a function of $N$. For each point, we calculate $r^*$ from 100 random datasets and we plot the sample mean and standard deviation. In the plotted example, $r_a,r_b,r_c = 1.1,1.4,3.5$.}
\label{fig:compare}
\end{center}
\vskip -0.2in
\end{figure}

For a less trivial example that highlights the shortcomings of other information-theoretic clustering criteria, we consider points drawn from the distribution in Fig.~\ref{fig:circle}. The clusters in this case are non-convex and unbalanced in size. Let $Y_r$ denote a cluster label that distinguishes whether a point is at radius greater than $r$ or not. We evaluate the quality of this partitioning as a function of $r$ and $N$ using CVR, the mutual information (MI) using the estimator in Eq.~\ref{eq:kraskov}, and the mutual information inspired objective NIC~\cite{faivishevsky}. For each objective, we label $r^* = \arg \operatorname{opt}_r \mbox{Objective}(Y_r,X)$. The results are shown in Fig.~\ref{fig:compare}. Note that the mutual information estimator we use is asymptotically unbiased~\cite{kraskov}, and so we expect a similar result for any mutual information estimator. We see that mutual information converges to an incorrect solution that equally splits the probability mass. Also note that while NIC is meant to approximate mutual information, it actually converges to a different incorrect value because it is not an unbiased estimator (see note in Sec.~\ref{sec:related}). We see that CVR performs best for all values of $N$ and converges quickly to the correct solution.  

\begin{figure}[ht]
\begin{center}
\includegraphics[width=0.88\columnwidth]{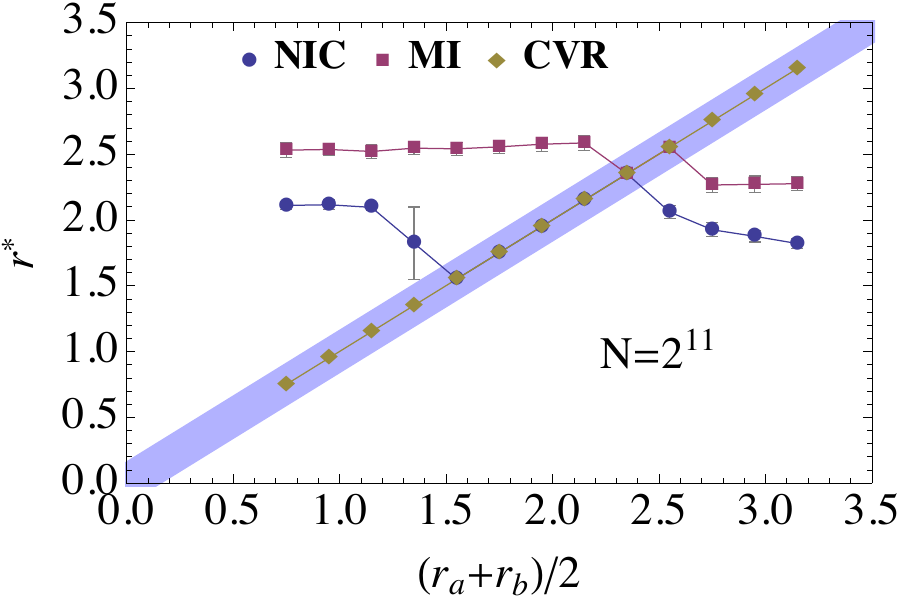}
\caption{We compare the accuracy of various information-theoretic clustering objectives over a range of parameter values using a fixed number of samples, $N=2^{11}$.  For the distribution in Fig.~\ref{fig:circle}, we set $r_c =3.5,r_b=r_a+0.3$ and we vary $r_a$. For each point, we calculate $r^*$ from 20 random datasets and we plot the sample mean and standard deviation. The shaded area indicates the correct range for $r^*$.}
\label{fig:comparer}
\end{center}
\end{figure}

Fig.~\ref{fig:comparer} shows that these results are robust to parameter changes, though NIC, unlike MI, converges to correct solutions if the cluster size imbalance is small. CVR is the only objective to prefer the correct partition over a wide range of parameter values.
While in this simple example, we only considered partitions defined by $Y_r$, we show in Sec.~\ref{sec:sdp} that our heuristic optimizer is able to search over all possible partitions to recover the correct, non-convex clusters.

\begin{figure}[ht] 
\centering
    \includegraphics[width=0.83\columnwidth]{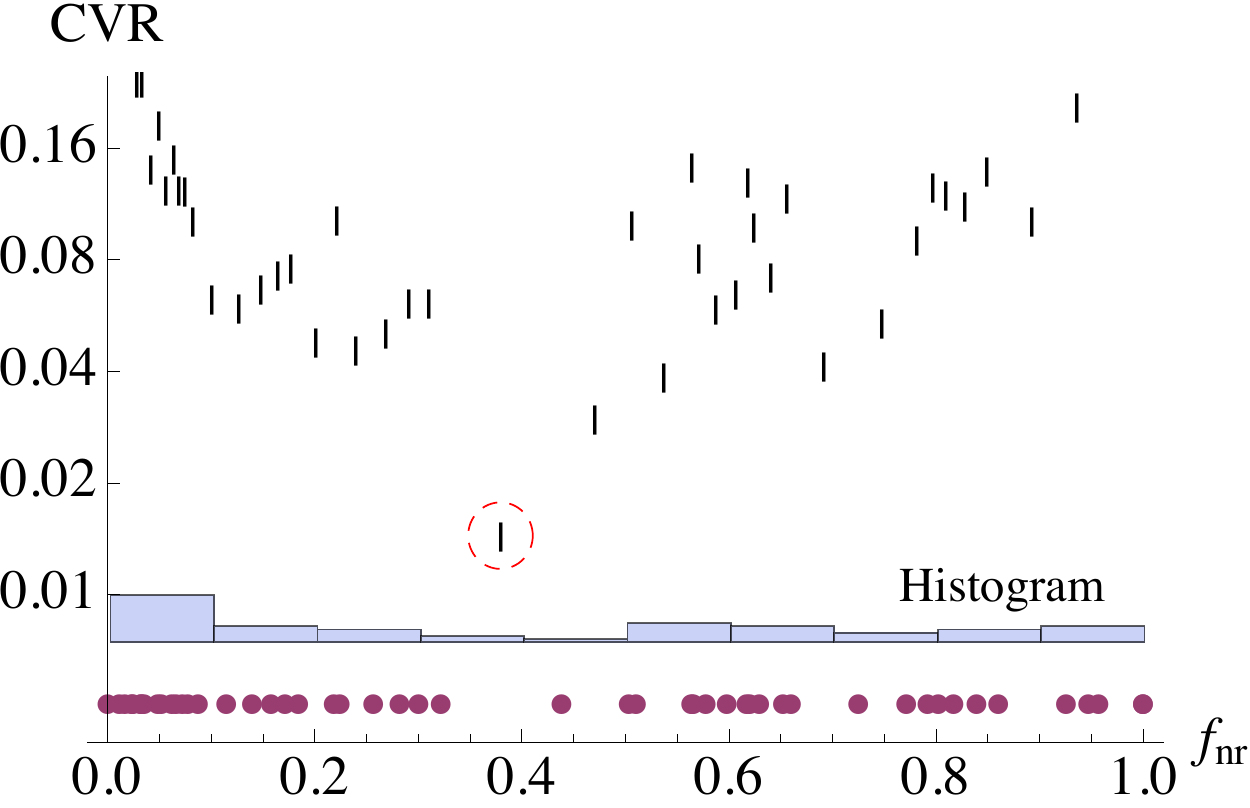} 
   \caption{We estimate CVR for editor behavior on Wikipedia.}
   \label{fig:cvtmore}
\vskip -0.2in
\end{figure}


\subsection{Real-world datasets}\label{sec:real}

\shortpara{Editor activity on Wikipedia} 
In Fig.~\ref{fig:cvtmore}, we consider the behavior of users who frequently edit the Wikipedia page for George W. Bush~\cite{dedeo-methods}. On Wikipedia, users may contest any article's point of view by directly editing the text. When a change is made by some user, any other user can choose to reject the change by ``reverting'' to a prior version. Certain types of users are likely to get into turf wars or, conversely, to engage in pro-social vandalism repair, so that a large fraction of their activity consists of ``reverts''. We look at the fraction of a user's activities that does not consist of reverting to previous versions, $f_{nr}$. For the 50 most active users on the George W. Bush Wikipedia page, we plot $f_{nr}$ in Fig.~\ref{fig:cvtmore}. Both CVR and Bayesian latent variable models like $k$-means discover the same natural cluster structure when used to determine a binary partition. 



\begin{table*}[th]
\begin{center}
  \begin{tabular}{c|ccc|ccc|cc} \hline
     &  &Stats&&  & Rand index &  && \\ \hline
  Dataset & $N$ &\#clusters&dim & k-means & ITC & CVR &CVR$_{bf}$&CVR$_{gt}$ \\ \hline
  \iris{} &150&3&4& 0.868&0.94&0.925&0.08&0.09 \\
  \wine{} &178&3&13& 0.927&0.92&0.936&0.18&0.25\\ 
  \glass{} &214&6&9& 0.678&0.75&0.671&0.33&1.16\\ \hline
  \end{tabular}
   \caption{Summary of results from clustering with the UCI datasets~\cite{uci}. The middle columns report cluster quality by calculating the Rand index~\cite{rand} of a candidate clustering with respect to the ground truth. For information-theoretic clustering (ITC), we report recent state-of-the-art results for optimizing mutual information using an SDP~\cite{feicluster}. The right columns report CVR for the ground truth clustering, $Y_{gt}$ and the best (lowest) CVR found by our heuristic, $Y_{bf}$.}  \label{fig:uci}%
   \end{center}
   \vskip -0.15in
\end{table*}


\shortpara{UCI datasets} We also consider three standard clustering datasets from the UCI Machine Learning database~\cite{uci}: \glass{}, \iris{}, and \wine{}. Although we cannot optimize the objective in Eq.~\ref{eq:cv} exactly, we describe details of a heuristic approach in Appendix~\ref{sec:sdp}. 
The results are summarized in Table~\ref{fig:uci}.  We achieve competitive results to $k$-means and the best of IT methods that have been previously compared~\cite{feicluster}.
Note that the cluster sizes for \iris{} and \glass{} are exactly and approximately balanced respectively which benefits mutual information since it is biased towards balanced solutions. 
 
 We also compare the CVR of the ground truth clustering to the best clustering found according to our heuristic. 
The CVR of the clusters found for \iris{} and \wine{} is comparable to that of the ground truth and much smaller than the maximal possible CVR of 1. On the other hand, the CVR for the ground truth clustering for \glass{} is actually larger than 1 (possible due to noisy estimators). This implies cluster labels of neighboring sample points are nearly random. Note that the Rand index~\cite{rand} of the clustering in which each point is its own cluster achieves a score of $0.74$, comparable to all the results listed in Table~\ref{fig:uci}. 

\section{Discussion}\label{sec:discussion}\label{sec:related}\label{sec:conclusion}

\paragraph{Related Work}

Many clustering approaches explicitly combine information-theoretic methods with a similarity measure~\cite{slonim,ric,gokcay}. That is they define some function $f(x^{(i)},x^{(j)})$ that characterizes the similarity of two points and then maximize intra-cluster similarity while simultaneously optimizing some information-theoretic quantity. While these methods side-step our critique of purely information-theoretic clustering methods, they also lose generality with their insistence on defining some {\em ad hoc} notions of similarity. 
A line of recent work has attempted to maximize the mutual information between data and cluster labels using non-parametric entropy estimators. 
Faivishevsky and Goldberger~\cite{faivishevsky} attempt to use a modified version of a non-parametric entropy estimator~\cite{kraskov} to create a more tractable objective for optimization.\footnote{It should be noted that the entropy estimator in~\cite{FaivishevskyG08}  is not asymptotically correct. They argue ``The MeanNN estimator exploits the fact that the kNN estimation is valid for every $k$ and therefore averaging estimators for all possible values of $k$ leads itself to a new estimator of the differential entropy''. The $k$-NN estimator is guaranteed to be consistent only when $k$ grows sufficiently slowly (sublinearly) with $N$, the number of samples; see~\cite{Wang2009} and references therein. Averaging all values of $k$ up to $N$ violates this condition. 
A simple example where their estimator can be seen to fail is the distribution in Fig.~\ref{fig:bar}(b). Regardless of $N$, their estimator is proportional to $\log L$, while the true entropy is independent of $L$. } 
Wang and Sha demonstrate a semidefinite optimization based on this criteria~\cite{feicluster}. 
Other attempts to define tractable optimization problems invoke a squared loss variant of mutual information~\cite{squaredloss} and estimators based on constructing minimum spanning trees~\cite{spanning}. 
While these methods are general and have shown some success, we showed in Sec.~\ref{sec:problem} and Sec.~\ref{sec:results} that these approaches are ultimately flawed and will fail in the large data limit.

We showed that our approach, unlike MI-based methods, is not biased towards balanced clusters. This is in keeping with our desire to minimize the number of assumptions.
On the other hand, there may be scenarios in which a bias towards balanced clusters is desirable. One line of work attempts to define intuitive properties of clustering methods (like sensitivity to cluster balance) and then categorize them with respect to these properties; see \cite{ackerman} and references therein.

\paragraph{Conclusion}
We have demonstrated why existing information-theoretic clustering methods are conceptually flawed and presented a principled solution without introducing any model-based assumptions\footnote{While we achieved our goal of avoiding reliance on a {\em global} similarity measure, non-parametric entropy estimators require a notion of adjacency to find data points that are nearest neighbors according to some metric. A major advantage of taking the information-theoretic approach is that $H(Y|X)$ is invariant under smooth, invertible transformations of the data ($X$). Therefore, any asymptotically unbiased estimator should converge to the same value regardless of the metric used for finding nearest neighbors.}. {\em Consistency under coarse-graining} is essential to the definition of entropy. Formally incorporating this notion in a finite-data setting provides a basic and foundational motivation for defining clusters. Alternately, our objective can be viewed as a way of minimizing the estimated uncertainty of cluster labels. 
Preliminary results indicate the feasibility of optimizing our criteria and also demonstrate competitive accuracy on standard benchmarks.

Information-theoretic learning methods are attractive because they make no assumptions about the underlying data while maintaining a clear, operational meaning. 
Real-world learning problems consist of finite samples of data from unknown distributions. 
To construct an information-theoretic foundation for unsupervised learning, we need to carefully refine our measures so that they make sense in finite-data regimes. 
We hope that further development of these ideas will contribute to that goal. 

\nocite{dhillon}


%

\subsubsection*{Acknowledgments}
{ We thank the reviewers for helpful comments. A.G. and G.V. were supported in part by the US AFOSR MURI grant FA9550-10-1-0569 and DTRA grant HDTRA1-10-1-0086. F.S. is supported by an Alfred P. Sloan Foundation
Research Fellowship. S.D. acknowledges the support of National Science Foundation grant EF-1137929.}
\clearpage

\bibliography{gversteeg,galstyan}
\bibliographystyle{icml2014}

\clearpage
\onecolumn
\section*{Supplementary Material}
\appendix
\section{Derivation of Eq.~\ref{eq:cv1}}\label{sec:derive}

We have samples $(\vx^\ii,y^\ii)$, with $\vx \in \mathbb R^d$ and $y \in \{1,2,\ldots,l\}$, and we would like to evaluate Eq.~\ref{eq:CV} or $\hat H(Y|X)$. 
We define this estimator in terms of existing entropy estimators, $\hat H(Y|X) \equiv \hat H(Y) +  \hat H (X|Y) - \hat H(X)$. 
The definition of the differential entropy estimators is given by Eq.~\ref{eq:kraskov}. 

To write down the standard plug-in estimator for discrete entropy, we first define $n_j \equiv \sum_{i=1}^N \delta_{y^\ii,j}$. 
$$\hat H(Y) = -\sum_{j=1}^l n_j/N \log (n_j/N)$$
Next we should unpack the second term, using the definition that $\bar \epsilon_{i,k}$ denotes the distance to the $k$-th nearest neighbor to point $x^\ii$ that is in the same cluster. 
\benn
\hat H (X|Y) &=& \sum_{j=1}^l p(y=j) \hat H(X|y=j) \\
&=& \sum_{j=1}^l n_j/N \left( \log (n_j/k)+c_{k,n_j} + \frac{d}{n_j}\sum_{i=1}^N \delta_{y^\ii,j} \log \bar \epsilon_{i,k} \right)
\eenn
We have used Eq.~\ref{eq:kraskov} in the second line. Next we expand and perform the sum over $j$ for the last term only, eliminating the delta function.
\benn
&=& \sum_{j=1}^l n_j/N \left( \log (n_j/N)+\log N/k +c_{k,n_j} \right) + \frac{d}{N}\sum_{i=1}^N  \log \bar \epsilon_{i,k} \\
&=& -\hat H(Y) +\log N/k + \sum_{j=1}^l c_{k,n_j} n_j/N + \frac{d}{N}\sum_{i=1}^N  \log \bar \epsilon_{i,k}
\eenn
Putting this all together and using the nearest neighbor for estimation, or $k=1$. 
\benn
\hat H(Y|X)  & = &  \hat H(Y) +  \hat H (X|Y) - \hat H(X) \\
&=& \log N + \sum_{j=1}^l  c_{1,n_j} n_j/N + \frac{d}{N}\sum_{i=1}^N  \log \bar \epsilon_{i,1} - \log N - c_{1,N} - \frac{d}{N}\sum_{i=1}^N \log  \epsilon_{i,1} \\
&=&  \left(\sum_{j=1}^l c_{1,n_j} n_j/N - c_{1,N} \right)  + \frac{d}{N}\sum_{i=1}^N  \log \frac{\bar \epsilon_{i,1}}{\epsilon_{i,1}}  \\
&\approx & \frac{d}{N}\sum_{i=1}^N  \log \frac{\bar \epsilon_{i,1}}{\epsilon_{i,1}} 
\eenn
The constant quickly goes to zero for reasonable values of $N$ and we neglect it.
Recalling the definition, $c_{k,N} =\psi(N)-\psi(k)+\log(2k/N)$. The constant term has the form 
$\sum_{j=1}^l (\psi(n_j)-\log(n_j)) n_j/N - (\psi(N)-\log(N)) $. Using a well-known expansion for the digamma function, $\psi(n) - \log(n) = -1/(2n) + O(1/(2n)^2)$. Applying this first order expansion to our expression gives a single term $(1-l)/N$.

There is an ambiguity in applying our entropy estimator in Eq.~\ref{eq:kraskov} to some subset of points defined by a cluster if there are only $k$ or fewer points in a cluster. Then we define, $\bar \epsilon_{i,k} =  \epsilon_{i,N-1}$ if $k > n_{y^\ii}+1$. This definition reflects the fact that our uncertainty is maximal if we are not given sufficient data. For instance, when $k=1$, that means we have a cluster that contains a single sample point. In that case, we estimate the entropy of this single point as maximal with respect to the full dataset. In principle, the maximum length scale could be set by other prior information, but we strive to make data-driven choices whenever possible. In practice, this choice penalizes very small clusters, and the details of the penalty makes little difference in the outcome.

\section{Heuristic Optimization}\label{sec:sdp}

The goal is to search for a natural coarse-graining, as defined in Eq.~\ref{eq:cv}. 
The number of ways of partitioning $N$ points into groups is very large and evaluating CV for any partitioning requires calculating all pairwise distances. Even calculating the change in CV from altering the cluster membership of a single point may require $O(N)$ operations. Finally, the figures in Sec.~\ref{sec:results} suggest that our optimization landscape is very rugged, so gradient-based methods are unlikely to succeed. Because of these difficulties, we suggest a heuristic optimization below. 

Our optimization proceeds first by generating a small number of candidate partitions that are likely to have small CV using a tractable semidefinite program.
Then we rank the candidate partitions according to CVR.
\benn
\hat H_T(Y|X) = \int_0^1 d\alpha~\hat H_\alpha (Y|X) = \frac{d}{N}  \sum_{i=1}^{N}\sum_{k=1}^{N-1} \frac{1}{k(k+1)} \log \frac{\bar \epsilon_{i,k}}{\epsilon_{i,k}},
\eenn
Clearly, the contribution from terms coming from $k$-th nearest neighbors quickly decreases with $k$. If we ignore $k \geq k_{max}$, then  
the objective is clearly minimized as long as the $k$-th nearest neighbors for point $x^\ii$ are all in the same cluster. We begin by relaxing our discrete cluster variable $y^\ii$ to be a continuous variable lying on a hypersphere, i.e., $\vy^\ii \in \mathbb R^{d'}$ and $\vy^\ii \cdot \vy^\ii =1$. If two points are close together in the $\vx$ space, we want them to be close together in $\vy$ space as well. We define a weighted adjacency matrix, $A_{i,j} = 1/(k(k+1))$ if $j$ is the $k$-th nearest neighbor to $i$.  
$$ \min_{|\vy^\ii|=1} \sum_{i,j} A_{i,j} (\vy^\ii - \vy^{(j)})^2 = \max_{|\vy^\ii|=1} \sum_{i,j} A_{i,j} \vy^\ii \cdot \vy^{(j)}$$
The optimal value of this optimization is for all $\vy^\ii$ to be equal. We need to add a term that forces all the $\vy^\ii$ as far apart as possible.
$$\max_{|\vy^\ii|=1} \sum_{i,j} (A_{i,j} - \beta) \vy^\ii \cdot \vy^{(j)}$$
We represent the Gram matrix with components $M_{ij} = \vy^\ii  \cdot \vy^{(j)}$, and define $\bar A_{i,j} = A_{i,j} - \beta$. Then our optimization takes the form:
\be\label{eq:sdp}
\max_{M \mbox{ \tiny is }p.s.d.} \mbox{Tr} \bar A \bullet M,
\ee
where p.s.d. denotes that $M$ is positive semidefinite, a necessary and sufficient condition for it to be a Gram matrix. 
This optimization is a semidefinite program and can be efficiently solved using convex optimization techniques. We used $k_{max} = 10$ and $\beta = 1/(k_{max}(k_{max}+1))$ and made no effort to optimize these parameters.

Once the optimal $M$ has been found, a discrete clustering can be found using the rounding method of Goemans and Williamson~\cite{maxcut}. First, taking the Cholesky decomposition of $M$ recovers the vectors $\vy^\ii$. Now to recover a discrete partition (into two groups) from these vectors we pick a random unit vector $\mathbf u$, and partition the data according to $y^\ii \equiv \mbox{sign} (\mathbf u \cdot \vy^\ii)$. We generated 200 candidate partitions this way.

Next, we calculate the CVR for each partition and pick the best (lowest) one. 
In the event we want multiple clusters, we combine the partition with the lowest CVR with each of the top 25 remaining candidate partitions and then we chose the one with the smallest overlap with the original partition (according to the Rand index). 
We continue this procedure until we have the desired number of clusters. 

\begin{SCfigure}[1.5][b]
\centering
\includegraphics[width=0.3\textwidth]{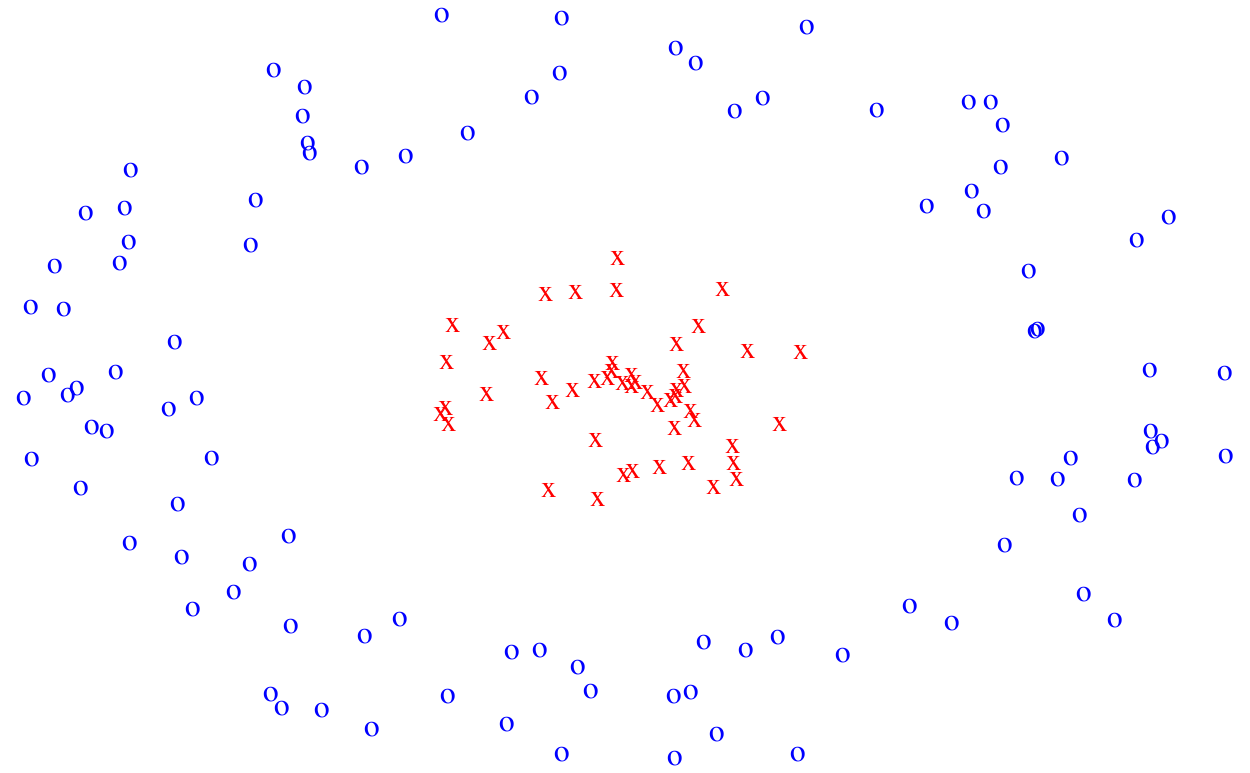}
\caption{An example of a non-convex clustering produced by our heuristic optimizer for Eq.~\ref{eq:cv}. The correct partition is found despite unbalanced cluster sizes.}
\label{fig:circle2}
\end{SCfigure} 

\end{document}